\documentclass[10pt,twocolumn,letterpaper]{article}

\usepackage{iccv}
\usepackage{times}
\usepackage{epsfig}
\usepackage{graphicx}
\usepackage{amsmath}
\usepackage{amssymb}

\usepackage[accsupp]{axessibility}
\usepackage{graphicx}
\usepackage{amsmath}
\usepackage{amssymb}
\usepackage{booktabs}

\usepackage[table]{xcolor}
\usepackage{color, colortbl}
\definecolor{LightCyan}{rgb}{0.88,1,1}

\usepackage[pagebackref=true,breaklinks=true,letterpaper=true,colorlinks,bookmarks=false]{hyperref}

\usepackage[T1]{fontenc}
\usepackage{graphicx}
\usepackage{latexsym}
\usepackage{xspace}
\usepackage{caption}
\usepackage{subcaption} 
\usepackage{cprotect}
\usepackage{hhline}
\usepackage{array}
\usepackage{float}
\usepackage{xcolor}
\usepackage[group-separator={,},output-decimal-marker = {.}, group-minimum-digits={3}]{siunitx}  
\usepackage{colortbl}
\usepackage{booktabs} 
\usepackage{multirow}
\usepackage[shortlabels]{enumitem} 
\usepackage{tablefootnote}
\usepackage{arydshln}

\definecolor{Gray}{gray}{0.6}

\usepackage{amssymb}
\usepackage{amsmath,amsfonts}
\usepackage{amsopn}
\usepackage{bm} 
\usepackage{multirow}
\newlength\savewidth

\newcommand{\ProbOpr}[1]{\mathbb{#1}}

\newcommand{\expect}[2]{%
\ifthenelse{\equal{#2}{}}{\ProbOpr{E}_{#1}}
{\ifthenelse{\equal{#1}{}}{\ProbOpr{E}\left[#2\right]}{\ProbOpr{E}_{#1}\left[#2\right]}}} 
\newcommand{\var}[2]{%
\ifthenelse{\equal{#2}{}}{\ProbOpr{VAR}_{#1}}
{\ifthenelse{\equal{#1}{}}{\ProbOpr{VAR}\left[#2\right]}{\ProbOpr{VAR}_{#1}\left[#2\right]}}} 

\newcommand{\eat}[1]{}

\newcommand{\mypartop}[1]{\vspace{0mm}\noindent\textbf{#1}.}
\newcommand{\mypar}[1]{\vspace{0.5em}\noindent\textbf{#1}.}

\newcommand{\vl}{{V\&L}\xspace}

\newcommand{\itform}{{$\langle image,text \rangle$}\xspace}
\newcommand{\icform}{{$\langle image,caption \rangle$}\xspace}

\newcommand{\baseline}{{\sc {NoPreSTU}}\xspace}

\newcommand{\ourmodel}{{\sc {PreSTU}}\xspace}
\newcommand{\ourmodelbf}{{\textbf{\textsc {PreSTU}}}\xspace}

\newcommand{\ourocr}{{\sc {ocr}}\xspace}

\newcommand{\oursplitocr}{{\sc {splitocr}}\xspace}
\newcommand{\oursplitocrbf}{{\textbf{\textsc {splitocr}}}\xspace}
\newcommand{\oursplitocrcap}{{\sc {splitocr\textrightarrow cap}}\xspace}

\newcommand{\oursplitocrvqa}{{\sc {splitocr\textrightarrow vqa}}\xspace}

\newcommand{\oursplitocrvqacap}{{\sc {splitocr\textrightarrow vqa/cap}}\xspace}

\newcommand{\ourcap}{{\sc {cap}}\xspace}
\newcommand{\ourcapbf}{{\textbf{\textsc {cap}}}\xspace}

\newcommand{\ourvqa}{{\sc {vqa}}\xspace}
\newcommand{\ourvqabf}{{\textbf{\textsc {vqa}}}\xspace}

\newcommand{\ourvqacap}{{\sc {vqa/cap}}\xspace}

\newcommand{\ourcapvqa}{{\sc {cap\textrightarrow vqa}}\xspace}
\newcommand{\ourcapsplitocr}{{\sc {cap\textrightarrow splitocr}}\xspace}

\newcommand{\oursplitocrcapvqa}{{\sc {splitocr\textrightarrow cap\textrightarrow vqa}}\xspace}

\newcommand{\mlmitmrpp}{{\sc {mlm+itm+rpp}}\xspace}
\newcommand{\mlm}{{\sc {mlm}}\xspace}
\newcommand{\itm}{{\sc {itm}}\xspace}
\newcommand{\rpp}{{\sc {rpp}}\xspace}

\newcommand{\lm}{{\sc {vlm}}\xspace}
\newcommand{\roilocal}{{\sc {roilocal}}\xspace}
\newcommand{\lmitmitc}{{\sc {vlm+itm+itc}}\xspace}
\newcommand{\itc}{{\sc {itc}}\xspace}

\newif\ifdraft
\drafttrue
\draftfalse
\ifdraft
  \newcommand{\beer}[1]{{\color{cyan}Beer: #1}\xspace}
  \newcommand{\jihyung}[1]{{\color{brown}Jihyung: #1}\xspace}
  \newcommand{\xichen}[1]{{\color{olive}Xi: #1}\xspace}
  \newcommand{\frank}[1]{{\color{orange}Frank: #1}\xspace}
  
  \newcommand{\seabass}[1]{{\color{magenta}Seabass: #1}\xspace}
  \newcommand{\radu}[1]{{\color{red}Radu: #1}\xspace}
\else
  \newcommand{\beer}[1]{}
  \newcommand{\jihyung}[1]{}
  \newcommand{\xichen}[1]{}
  \newcommand{\frank}[1]{}
  \newcommand{\harry}[1]{}
  \newcommand{\seabass}[1]{}
  \newcommand{\radu}[1]{}
\fi

\usepackage[capitalize]{cleveref}
\crefname{section}{Sec.}{Secs.}
\Crefname{section}{Section}{Sections}
\Crefname{table}{Table}{Tables}
\crefname{table}{Tab.}{Tabs.}

\iccvfinalcopy 

\def\iccvPaperID{11076} 

\ificcvfinal\fi

\begin{document}

\title{PreSTU: Pre-Training for 
Scene-Text Understanding}

\author{Jihyung Kil$^1$\thanks{\quad Work done at Google Research.}$\footnotemark[1]$ \quad Soravit Changpinyo$^2$ \quad \\
{Xi Chen$^2$} \quad {Hexiang Hu$^2$} \quad {Sebastian Goodman$^2$} \quad {Wei-Lun Chao$^1$} \quad {Radu Soricut$^2$} \\
$^1$The Ohio State University \quad $^2$Google Research \\
{\tt\small\{kil.5,chao.209\}@osu.edu} \\ {\tt\small\{schangpi,chillxichen,hexiang,seabass,rsoricut\}@google.com}}

\maketitle
\ificcvfinal\thispagestyle{empty}\fi

\begin{abstract}
The ability to recognize and reason about text embedded in visual inputs is often lacking in vision-and-language (\vl) models, perhaps because \vl pre-training methods have often failed to include such an ability in their training objective. In this paper, we propose \ourmodel, a novel pre-training recipe dedicated to scene-text understanding (STU). \ourmodel introduces OCR-aware pre-training objectives that encourage the model to recognize text from an image and connect it to the rest of the image content. We implement \ourmodel using a simple transformer-based encoder-decoder architecture, combined with large-scale image-text datasets with scene text obtained from an off-the-shelf OCR system. We empirically demonstrate the effectiveness of this pre-training approach on eight visual question answering and four image captioning benchmarks.
\end{abstract}
\section{Introduction}
\label{sec:intro}

\begin{figure}
    \centering
    \centerline{\includegraphics[width=\linewidth]{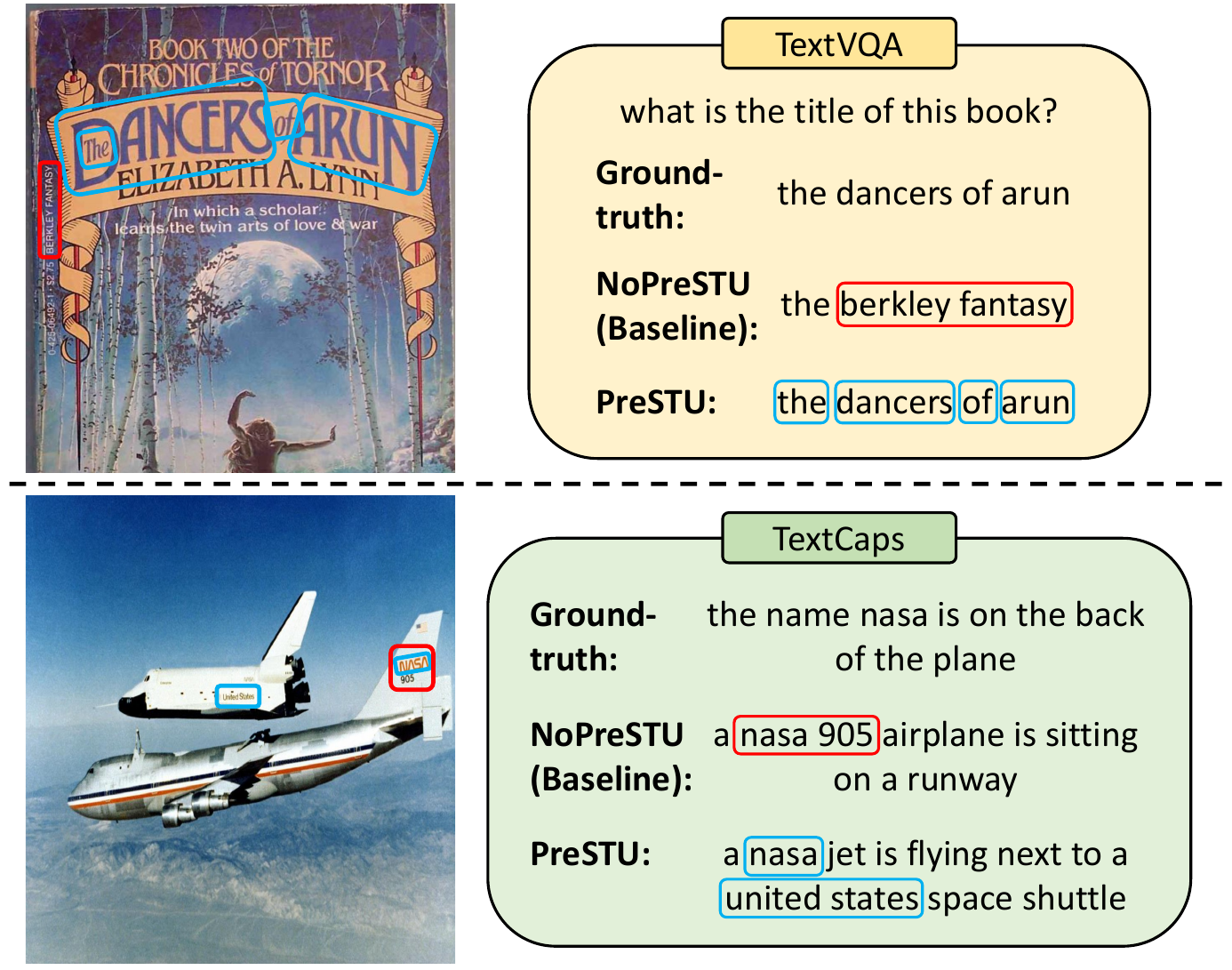}}
    \vspace{-5pt}
    \caption{\textbf{Example of scene-text understanding (STU) tasks.}
    \baseline (baseline) and \ourmodel share the same \vl model, but \ourmodel is pre-trained on our proposed pre-training objectives.
    Scene texts are highlighted by bounding boxes.
    Unlike the baseline, \ourmodel correctly predicts the title of the book on scene-text VQA (TextVQA~\cite{textvqa}) and even generates a more detailed scene-text caption (\eg, ``united states space shuttle'') than the ground-truth annotated by humans (TextCaps~\cite{textcaps}).}
    \label{fig:stu_task}
    \vspace{-15pt}
\end{figure}

Understanding the role of text as it appears in the context of a visual scene is important in various real-world applications, \eg, from automatically organizing images of receipts, to assisting visually-impaired users in overcoming challenges related to comprehension of non-Braille writing in their surroundings, to enabling autonomous robots to make safe decisions in environments designed for humans.
As a result, scene-text understanding (STU) has received increased attention in vision-and-language (\vl) understanding tasks, such as visual question answering (VQA) \cite{textvqa,stvqa,mishra2019ocr,wang2020general,docvqa,infographicvqa,chartqa} or image captioning~\cite{textcaps,vizwizcap,widgetcaptioning}. Please see \autoref{fig:stu_task} for an illustration.

We identify two distinct capabilities that models targeting STU must address:
(i) \emph{recognizing} text in a visual scene
and (ii) \emph{connecting} the text to its context in the scene.
Previous solutions that target STU tasks~\cite{textvqa,textcaps,m4c,tap} often delegate scene-text recognition to off-the-shelf OCR (Optical Character Recognition) systems~\cite{textcaps,borisyuk2018rosetta} and model the visual context using pre-computed object-detection features.
These two streams of information (noisy OCR strings and visual features on detected objects) are used as input into a \vl{} model.
While achieving decent results, these methods heavily rely on the quality of the upstream OCR system and lack a direct connection between the text being recognized and a high-fidelity representation of its context.

More concretely, previous methods have not fully explored pre-training objectives that specifically target STU.
In general, \vl{} pre-training objectives (\eg, masked language modeling, image-text matching \cite{lu19vilbert}, etc.) have been proven effective for learning and became the go-to approach in \vl{} research.
However, these objectives typically do not require a model to understand the role of text embedded in a visual context.
For instance, LaTr~\cite{latr} ignores the visual context during pre-training and instead focuses on modeling the co-occurrence statistics of layout-aware text-only OCR tokens.
Even in systems that do perform STU pre-training, such as TAP~\cite{tap}, their models are built upon the aforementioned pipeline.
Specifically, TAP represents the visual input by a set of object features detected and extracted by FRCNN~\cite{fasterrcnn}. As a result, it may lose some visual contexts that cannot be captured by objectness (\eg, activities) but are relevant to understand the role of recognized text.

\beer{I think we should talk about LaTr here to start inform the reader that 1) we are aware of it, 2) we are different, highlighting that their pre-training objectives do not even consider the visual context and leverage the co-occurance statistics of OCR text tokens.}

In this paper, we address such a challenge by incorporating an OCR-aware learning objective in the context of a high-fidelity representation of the image context.
We adopt a Transformer-based~\cite{transformers} encoder-decoder \vl architecture, using a T5~\cite{t5} backbone.
The model takes both image and text inputs. For the former, we extract fine-tunable visual features directly from image \emph{pixels} using a ViT~\cite{vit} encoder, rather than adopting frozen visual features from pre-detected objects~\cite{fasterrcnn}.
For the latter, we concatenate task-specific text tokens (\eg, task prompts) with tokens extracted from an off-the-shelf OCR system, in a manner that allows the model to interpret (via the prompt) the OCR tokens in the context of the image.

\begin{figure*}
    \centering
    \centerline{\includegraphics[width=0.95\linewidth]{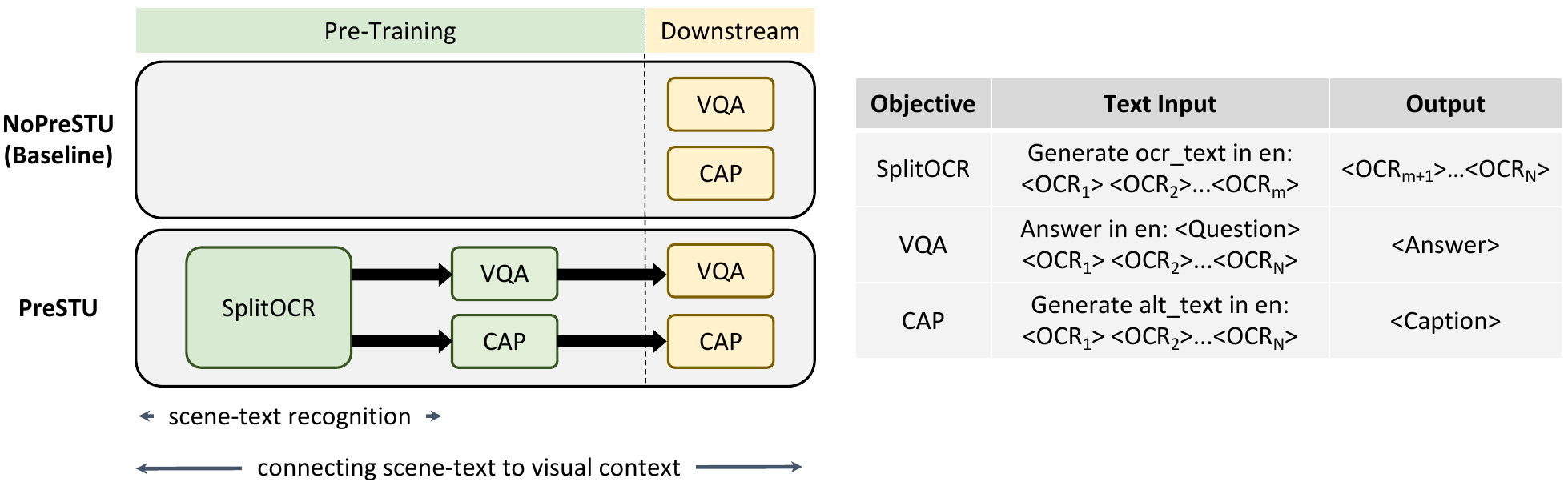}}
    \captionsetup{width=.95\textwidth}
    \vspace{-5pt}
    \caption{\textbf{Our proposed pipeline.} Left: Comparison between \ourmodel and \baseline (baseline) we want to compare against. Green denotes the \ourmodel pre-training phase and yellow the downstream/fine-tuning phase. \oursplitocr encourages scene-text recognition as well as the learning of the connection between scene text and its visual context; \ourvqa and \ourcap further strengthen that connection. Right: The text input and output for each objective. All objectives utilize OCR signals. See \autoref{fig:arch_prestu} for the architecture of \ourmodel.}
    \label{fig:approach}
    \vspace{-10pt}
\end{figure*}

Building upon this model, we propose \ourmodel, a novel recipe for \textbf{Pre}-training for \textbf{S}cene-\textbf{T}ext \textbf{U}nderstanding  (\autoref{fig:approach}). \ourmodel consists of two main steps.
First, it teaches the model to recognize scene text from image pixels\footnote{This makes our model more robust to the quality of OCR
systems.} and at the same time connect scene text to the visual context.
Specifically, given an image and the ``part'' of the scene texts in the image, the model is pre-trained to predict the ``rest'' of the scene texts. We call this step \oursplitocr. Second, it teaches the model to further strengthen the connection between scene text and visual context by pre-training with OCR-aware downstream tasks (\eg, \ourvqa and \ourcap). 
For pre-training, we leverage large-scale image-text resources \cite{cc3m,cc12m,stvqa}, with the (noisy) scene text extracted by the off-the-shelf OCR system (Google Cloud OCR\footnote{https://cloud.google.com/vision/docs/ocr}).

We validate \ourmodel on eight VQA (ST-VQA~\cite{stvqa}, TextVQA~\cite{textvqa}, VizWiz-VQA~\cite{vizwiz}, VQAv2~\cite{vqa2}, OCR-VQA~\cite{mishra2019ocr}, DocVQA~\cite{docvqa}, ChartQA~\cite{chartqa}, AI2D~\cite{AI2D}) and four image captioning (TextCaps~\cite{textcaps}, VizWiz-Captions~\cite{vizwizcap}, WidgetCap~\cite{widgetcaptioning}, Screen2Words~\cite{screen2words}) benchmarks. Our OCR-aware objectives \oursplitocr, \ourvqa, and \ourcap  are significantly beneficial. For instance, compared with strong baselines which take OCR signals as input, we observe more than 10\% absolute gain on TextVQA and 42 CIDEr point gains on TextCaps (\autoref{fig:stu_task}). Finally, we conduct comprehensive experiments to understand which factors contribute to effective STU pre-training. In summary, our contributions are as follows:
\beer{Should we highlight in the contributions somewhere that we also show that our objective is also useful in the absence of OCR signals during downstream task evaluation.}
\begin{itemize} [itemsep=0pt,topsep=0.0pt,leftmargin=10pt,partopsep=0pt]
    \item We propose \ourmodel, a simple and effective pre-training recipe with OCR-aware objectives designed for scene-text understanding (\S\ref{sec:approach}).
    \item We show that our objectives consistently lead to improved scene-text understanding on twelve diverse downstream VQA / image captioning tasks (\S\ref{ssec:exp_res}) and even on cases when OCR signals are absent during downstream tasks (\S\ref{ssec:analysis}).
    \item We perform detailed analyses to understand the effect of our design choices on STU performance (\S\ref{ssec:analysis}).
\end{itemize}

\section{PreSTU: Pre-Training for Scene-Text Understanding}
\label{sec:approach}

\begin{figure}
    \centering
    \centerline{\includegraphics[width=0.8\linewidth]{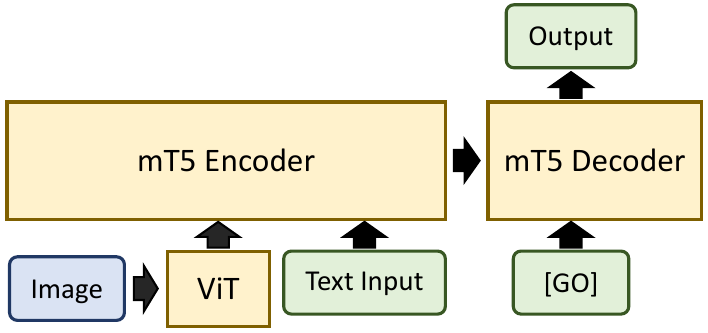}}
    \caption{\textbf{\vl model architecture used in all of our experiments.} We use a simple transformer-based encoder-decoder (pre-trained ViT~\cite{vit} + mT5~\cite{mt5}) transforming image and text inputs to the text output. Green box: text input/output. Blue box: visual input. Yellow box: model blocks. See \autoref{fig:approach} for the input-output pairs for different objectives.}
    \label{fig:arch_prestu}
\vspace{-15pt}
\end{figure}

\autoref{fig:approach} provides an overview of \ourmodel OCR-aware objectives and their input-output format.
In what follows, we first describe our starting point: model architecture and OCR signals (\S\ref{setup}). Then, we describe our recipe for pre-training (\S\ref{pretrain_stage}), including the objectives, \oursplitocrbf, \ourvqabf, and \ourcapbf (\S\ref{objectives}), and data sources (\S\ref{data}). Finally, we describe the fine-tuning stage and target benchmarks (\S\ref{finetune_stage}).

\subsection{Setup}
\label{setup}

\mypartop{\vl model architecture}
Our main architecture is illustrated in \autoref{fig:arch_prestu}. We start from an encoder-decoder \vl architecture which unifies image-to-text (e.g., image captioning) and image+text-to-text (e.g., VQA) tasks.
The pre-trained vision encoder is ViT-B/16~\cite{vit}, and the pre-trained language encoder-decoder is mT5-Base~\cite{mt5}. Specifically, ViT is a transformer-based encoder that takes a sequence of image patches as input, pre-trained on an image classification task. mT5 is a multilingual variant of text-to-text transformers T5~\cite{t5}, pre-trained on a massive multilingual text corpus with the span corruption objective. See more details in the supplementary material. 

As mentioned in LaTr~\cite{latr}, this starting point leads to modeling advantages over existing model architectures for STU tasks.
First, we believe that understanding the role of OCR text in the visual context is much easier from image pixels, making ViT a natural choice.
Second, mT5 uses wordpiece vocab to encode and decode text tokens; thus a certain level of robustness to the noise in the input OCR texts comes with it by default. On the other hand, M4C~\cite{m4c} and TAP~\cite{tap} resort to a more complicated solution of using fastText~\cite{fasttext} and Pyramidal Histogram of Characters features~\cite{phoc}.
Third, mT5 is an encoder-decoder model which enables to generate the open-ended text. 
This is suitable for general image captioning and scene-text VQA where the answers tend to be out-of-vocab. 
In contrast, most prior works~\cite{textvqa,m4c,tap,wang2022tag,logos} treat VQA as answer vocab-based classification. 
Lastly, our model is built upon well-developed vanilla unimodal building blocks in vision and NLP. We deliberately choose this general encoder-decoder architecture to push for the applicability of our objectives. Such a design choice allows us to develop less model-dependent pre-training objectives.

\mypar{Image resolution} Unless stated otherwise, we use the image resolution of 640x640 in all of our experiments.

\mypar{OCR signals} We obtain OCR signals from Google Cloud OCR for all pre-training and downstream datasets in our experiments. They come in the form of a set of texts and their corresponding box coordinates in the image (\ie, object detection-like). We order OCR texts based on their locations, top-left to bottom-right and concatenate them with the T5 separator \textsc{</s>}. 
This allows models to implicitly learn the scene text's spatial information and standarize the target output sequence during training.
Unless stated otherwise, we use these sorted \emph{silver} OCR texts in all of our experiments.

\subsection{Pre-Training Stage}
\label{pretrain_stage}

\subsubsection{PreSTU Objectives}
\label{objectives}

We consider two sets of OCR-aware pre-training objectives for scene-text understanding.

\mypar{Task-agnostic objective: SplitOCR} Inspired by the impressive performance of the visual language modeling pre-training objective for image+text-to-text downstream tasks~\cite{wang2022simvlm}, we propose an OCR-aware pre-training objective called \oursplitocr. 
This objective is designed to be downstream task-agnostic, focusing on teaching the two core capabilities for STU: recognizing scene text and connecting it to the visual context.

We randomly split the OCR texts into two parts and use the first part as additional input and the second part as a target. Recall that we have ordered the OCR texts based on their locations such that the model can recognize them in a consistent manner. Note that if the splitting point is right at the beginning of the OCR sequence, the model performs a simplified version of the traditional Optical Character Recognition task (\ie, predicting the whole OCR tokens).
We denote this by \ourocr in \autoref{tab:all_cases} and also compare it with \oursplitocr in our ablation studies.

\textbf{Why \oursplitocr?} 
\oursplitocr equips the model with the abilities to recognize scene text and connect it to the visual context in a unified, seamless manner. Specifically, operating \oursplitocr upon the {\color{magenta}``first part''} of OCR tokens and the image pixels (not pre-extracted global or object detection features) and predicting the {\color{cyan}``second part''} of OCR tokens requires the model to (i) identify which {\color{cyan}scene text} in the image \emph{still} needs to be recognized, inherently connecting the input {\color{magenta}scene text} to its visual context; (ii) perform the OCR \emph{task}, inherently acquiring the scene-text recognition skill.

\mypar{Task-specific objectives: VQA and CAP}
We propose OCR-aware downstream-task-specific pre-training objectives on top of \oursplitocr{}. We consider two objectives based on our downstream tasks: (i) VQA which predicts the target answer from the question prompt, the visual question, and OCR texts and (ii) CAP which predicts the target caption from the caption prompt and OCR texts. This is similar to previous approaches to STU, except that we encode the image pixels, not features from pre-detected regions.

\textbf{Why VQA or CAP?} Task-specific objectives aim to achieve two goals. First, they further encourage the learning of the relationship between scene text and its visual context through direct interaction between input image pixels and input OCR texts. Second, it eases the knowledge transfer from pre-training to fine-tuning since task-specific objectives share the same input format as that of the downstream tasks (\S\ref{finetune_stage}).
See \autoref{fig:approach} for more details.

\subsubsection{Pre-Training Data}
\label{data}
Our main pre-training data is CC15M, the union of two popular image-text datasets: Conceptual Captions (CC3M)~\cite{cc3m} and Conceptual 12M (CC12M)~\cite{cc12m}.\footnote{Due to expired URLs, only 13M \icform pairs are used in our experiments.} CC3M consists of 3.3M \icform pairs, obtained by processing raw alt-text descriptions from the Web. CC12M extends CC3M by relaxing its over-restrictive filtering pipeline. We use CC15M for \oursplitocr{} and \ourcap{} pre-training. Note that the captions of CC15M are not used for \oursplitocr{} and their images are not necessarily scene text-related. See more details in the supplementary material.

Since CC15M does not have data in the form of visual questions and their answers for us to leverage, we resort to ST-VQA~\cite{stvqa}. 
It is a scene-text VQA dataset whose images are collected from 6 diverse data sources (COCO-Text~\cite{veit2016coco}, Visual Genome~\cite{krishna2017visual}, VizWiz~\cite{vizwiz}, ICDAR~\cite{karatzas2013icdar,karatzas2015icdar}, ImageNet~\cite{deng2009imagenet}, IIIT-STR~\cite{mishra2013image}). 
We use its training set for pre-training.
We use ST-VQA as pre-training data for other VQA benchmarks as well as a downstream benchmark for testing \oursplitocr{} (\S\ref{finetune_stage}).

\subsection{Fine-tuning Stage}
\label{finetune_stage}
In all of our downstream scene-text \vl tasks, the input-output pairs follow the same format as either VQA or CAP ( with OCR text tokens as input.) The only difference from the task-specific pre-training is the training data.

We validate \ourmodel on twelve datasets related to VQA and image captioning tasks.
ST-VQA, TextVQA, and TextCaps are the main benchmarks for STU. We also consider other scene-text domains, including book (OCR-VQA), document (DocVQA), illustration (ChartQA), diagram (AI2D), and screenshot domains (WidgetCap and Screen2Words).
VizWiz-VQA and VizWiz-Captions are for the blind and heavily involve STU. 
VQAv2 is a general VQA dataset.
See complete details in the supplementary material.

\subsection{Discussion}
\label{approach_discuss}
We compare \ourmodel with two well-known prior STU works TAP~\cite{tap} and LaTr~\cite{latr}. 
In terms of modeling, TAP leverages two conventional \vl objectives: visual-region masked language modeling and image-text matching, as well as the objective of learning the relative spatial position of two OCR text detections. TAP models the image using object-based features~\cite{fasterrcnn}, which we believe is a suboptimal visual context. Besides, TAP adopts vocab-based classification, less suitable for some STU tasks which are full of out-of-vocab words. LaTr overcomes those weaknesses by adopting a similar \vl architecture to ours (ViT-B/16 / T5\textsubscript{large}). However, its pre-training objective does not involve the visual component (ViT). Instead, it only pre-trains its language component to learn the co-occurrence statistics of layout-aware OCR tokens. As the visual component is distorted or absent during pre-training, these models do not inherently learn the two essential STU capabilities, and would likely suffer in a case when OCR signals are absent during downstream tasks.
In contrast, \ourmodel fully embraces the visual component. As shown in \S\ref{ssec:analysis}, this brings a huge benefit especially when OCR signals are not available. See a more detailed comparison in \S\ref{exp_sota}.

In terms of pre-training data, TAP aggregates scene-text \emph{dedicated} downstream data, including ST-VQA, TextVQA, TextCaps, and OCR-CC. Thus, while it aligns well with the corresponding downstream tasks, it is less generalizable to other \vl tasks. In contrast, \ourmodel adopts \emph{general} pre-training data (\ie, CC15M), providing a more flexible interface for \vl tasks. Besides, LaTr argues that pre-training on document images is a better choice since acquiring large quantities of natural images with scene text for pre-training is challenging and hard to scale, and the amount of text is often sparse. Our work challenges this assumption and shows that one can pre-train effectively for STU on natural images with minimal preprocessing. (\ie, nothing beyond extracting OCR signals).

Finally, in terms of evaluation as we will show next, our experiments are done on a much wider range of benchmarks than before. This is in stark contrast to existing works which often focus on three benchmarks at most.

\begin{table*}[t]
\centering
\small
\tabcolsep 5.5pt
\renewcommand\arraystretch{1.0}
\begin{tabular}{cccccc}
\toprule
\multirow{2}[3]{*}{Model} &
\multirow{2}{*}{Pre-training} &
\multicolumn{4}{c}{Test Benchmark} \\
\cmidrule(r){3-6}
& Objective & ST-VQA & TextVQA & VizWiz-VQA & VQAv2 \\
& & ANLS & Acc & Acc & Acc \\
\midrule
\baseline & - & 56.7 & 44.8 & 57.7 / 57.2 & 74.8 / 75.2 \\
\midrule
\multirow{3}{*}{{\rotatebox[origin=c]{0}{\ourmodel}}}
& \ourvqa & N/A & 48.3 & 58.3 / 57.6 & 75.0 / 75.0 \\
& \oursplitocr & \textbf{65.5} & 55.2 & 61.9 / 61.3 & \textbf{76.0} / \textbf{76.2} \\
& \oursplitocrvqa & N/A & \textbf{56.3} & \textbf{62.5} / \textbf{62.0} & \textbf{76.1} / \textbf{76.1} \\
\bottomrule
\end{tabular}
\vspace{-5pt}
\captionsetup{width=0.95\textwidth} 
\caption{\small \textbf{Effectiveness of \ourmodelbf objectives on VQA.} Our pre-training objectives (\ourvqa, \oursplitocr, \oursplitocrvqa) show consistent gains over the baseline on all VQA benchmarks. We use CC15M for \oursplitocr pre-training and ST-VQA for \ourvqa pre-training. Since ST-VQA for \ourvqa pre-training, we mark \ourvqa and \oursplitocrvqa as ``N/A''. Results are reported on the test set for ST-VQA, test-std for TextVQA, and test-dev/test-std for VizWiz-VQA and VQAv2.}
\label{tab:main_vqa}
\end{table*}

\begin{table*}[t]
\centering
\small
\renewcommand\arraystretch{1.0}
\begin{tabular}{cccccccccccc}
\toprule
\multirow{2}[2]{*}{Model} &
\multirow{2}{*}{Pre-training} &
\multicolumn{5}{c}{TextCaps test-std} &
\multicolumn{5}{c}{VizWiz-Captions test-std} \\
\cmidrule(r){3-7} \cmidrule(r){8-12}
& Objective & B & M & R & S & C & B & M & R & S & C \\
\midrule
\baseline & - & 23.4 & 21.0 & 45.0 & 13.6 & 96.9 & 29.4 & 22.6 & 49.9 & 18.5 & 87.2 \\
\midrule
\multirow{3}[2]{*}{{\rotatebox[origin=c]{0}{\ourmodel}}} 
& \ourcap & 31.6 & 25.6 & 51.5 & 18.7 & 133.1 & 33.7 & 24.5 & 52.8 & 20.8 & 103.1 \\
& \oursplitocr & 28.5 & 23.9 & 48.9 & 16.3 & 126.1 & 29.8 & 22.6 & 50.3 & 18.6 & 90.2 \\
& \oursplitocrcap & \textbf{32.8} & \textbf{26.2} & \textbf{52.2} & \textbf{19.1} & \textbf{139.1} & \textbf{34.3} & \textbf{24.7} & \textbf{53.4} & \textbf{21.1} & \textbf{105.6} \\
\bottomrule
\end{tabular}
\captionsetup{width=0.95\textwidth}
\vspace{-5pt}
\caption{\small \textbf{Effectiveness of \ourmodelbf objectives on image captioning.} Our pre-training objectives (\ourcap, \oursplitocr, \oursplitocrcap) show significant gains over the baseline on all image captioning benchmarks, with \oursplitocrcap performing best. We use CC15M for both \oursplitocr and \ourcap pre-training. B: BLEU@4, M: METEOR, R: ROUGE-L, S: SPICE, C: CIDEr.}
\vspace{-10pt}
\label{tab:main_cap}
\end{table*}
\section{Experimental Results}
\label{sec:exp}

\mypar{Baselines}
We denote by \baseline our main baseline. It is the same pre-trained \vl model as \ourmodel (\ie, ViT-B/16 / mT5) but \emph{not} pre-trained with any of our pre-training objectives.

\mypar{Metrics}
For VQA tasks, we use standard VQA accuracy following~\cite{textvqa,tap,git}. It is the average score over nine subsets of the ground-truth ten answers, where each score is:
$min(\frac{\# answer\ occurrences}{3}, 1)$. 
For ST-VQA/DocVQA, we use Average Normalized Levenshtein Similarity (ANLS), \emph{softly} penalizing the model's mistakes on scene-text recognition. For ChartQA, we report its official metric, a relaxed accuracy that allows a minor inaccuracy for numeric answers.
For image captioning tasks, we use their standard evaluation metrics, including BLEU~\cite{papineni2002bleu}, METEOR~\cite{denkowski2014meteor}, ROUGE-L~\cite{lin2004rouge}, SPICE~\cite{anderson2016spice}, and CIDEr~\cite{cider}. 

\subsection{Main Results}
\label{ssec:exp_res}

The main goal of our experiments is to assess the utility of our pre-training objectives \oursplitocr and \ourvqa/\ourcap in VQA (\S\ref{vqa}) and image captioning (\S\ref{ic}) tasks. 

\subsubsection{VQA}
\label{vqa}
\autoref{tab:main_vqa} summarizes our main results on VQA tasks, including ST-VQA, TextVQA, VizWiz-VQA, and VQAv2.
\oursplitocr{} outperforms the baseline (\ie, without our STU pre-training) by a large margin on scene-text-heavy VQA tasks, more than +8.8 ANLS on ST-VQA, +10.4\% on TextVQA, and +4.1\% on VizWiz-VQA. With \oursplitocrvqa{}, we slightly but significantly improve the performance further on TextVQA and VizWiz-VQA, +1.1\% and 0.7\%, respectively. 
These results show the utility and applicability of our pre-training objectives for improving scene-text understanding.

\oursplitocr{} and \ourvqa{} are complementary on scene-text-heavy VQA tasks (TextVQA/VizWiz-VQA), where each of them alone underperforms \oursplitocrvqa{}. Additionally, we observe the first-stage pre-training via \oursplitocr{} is more beneficial than the second-stage task-specific pre-training \ourvqa. This could be due to the superiority of \oursplitocr{} or the lack of large-scale scene-text VQA pre-training data, or both.
We identify data development for scene-text VQA as an open research question.

Our results also highlight the importance of STU in general real-world VQA (\ie, not specially designed for STU). 
We observe a slight but significant improvement over the baseline on VQAv2 and a more significant improvement on VizWiz-VQA for blind people. 
We attribute this to a subset of questions that require text recognition and reasoning skills~\cite{zeng2020vision}.
We believe this is an important step since these questions are considered ``hard to learn'' or even ``outliers'' that work against VQA algorithms~\cite{swayamdipta2020dataset,karamcheti2021mind}.

\subsubsection{Image Captioning}
\label{ic}
\autoref{tab:main_cap} summarizes our main results on image captioning tasks, TextCaps and VizWiz-Captions.
Aligned with the VQA results, \oursplitocr{} significantly improves over the baseline across all evaluation metrics, with \oursplitocrcap{} performing best. The gain is notably 42.2 CIDEr points on TextCaps, and 18.4 on VizWiz-Captions. Overall, we highlight the usefulness of \oursplitocr{} across \vl{} tasks with different input-output formats. 

Similar to the VQA results, \oursplitocr{} and \ourcap{} are complementary. However, \ourcap{} alone is more beneficial than \oursplitocr{} alone. We attribute this to our large-scale web-based image-text data that is already suitable for \ourcap{} pre-training. Despite such a strong \ourcap{} model, \oursplitocr{} still provides an additional benefit.


\subsubsection{Applicability to Other Scene-Text Domains}
\label{other_stu_domain}
 Unlike prior STU literature~\cite{tap, wang2022tag, logos, latr, ConCap, UniTNT}, we further explore other scene-text domains (\autoref{tab:other_domains}). We show that PreSTU is also effective on book (OCR-VQA), document (DocVQA), illustration (ChartQA), diagram (AI2D), and screenshot domains (WidgetCap \& Screen2Words). This demonstrates the applicability of \ourmodel to many different real-world STU problems. 

\begin{table}[t]
\centering
\footnotesize
\tabcolsep 1pt
\begin{tabular}{ccccccc}
\toprule
\multirow{3}{*}{Model} & OCR & Doc & Chart & \multirow{2}{*}{AI2D} & Widget & Screen2 \\
& VQA & VQA & QA & & Cap & Words \\
 & \%Acc & \%ANLS & \%RelaxedAcc & \%Acc & CIDEr & CIDEr  \\
\midrule
NoPreSTU & 71.5 & 47.5 & 40.5 & 64.5 & 63.9 & 98.5 \\
\midrule
PreSTU-SplitOCR & 72.2 & 50.1 & 50.7 & 69.3 & 125.6 & 113.8\\
\bottomrule
\end{tabular}
\vspace{-8pt}
\caption{\small \textbf{PreSTU on other scene-text domains (Val split).} See \S\ref{other_stu_domain} for a detailed discussion.}
\label{tab:other_domains}
\vspace{-15pt}
\end{table}

\subsubsection{Comparison to Prior Works}
\label{exp_sota}
\jihyung{Need to cut this section}
\beer{I feel we should add something along this line at the beginning of this section to make sure the reader understand its purpose since this is the main complaint from CVPR reviews.}
So far our results provide strong evidence for the benefit of our proposed objectives. In this section, we provide a comparison to prior works as further context. While apples-to-apples comparison has become increasingly difficult, we make our best attempt to analyze our results in the context of these works. For example, TAP's objective has coupled the use of object detection signals, which we do not resort to. More importantly, many prior works~\cite{latr,flamingo,git} do not release code, rely on private data, and/or require too large-scale pre-training that is prohibitively costly to reproduce.

We first compare \ourmodel to recent works focusing on STU tasks (Rows Non-TAP to LaTr in \autoref{tab:comp_sota}). Overall, \ourmodel establishes strong results on all tasks. Concretely, PreSTU achieves better results than all prior smaller-scale works (\ie, TAP, TAG, LOGOS). More interestingly, with much less data, we even outperform two larger models ConCap/UniTNT (139.1 vs.~105.6/109.4 in CIDEr) on TextCaps and (56.3\% vs.~55.4\%) on TextVQA. 

PreSTU, however, performs worse than another larger model LaTr on TextVQA/ST-VQA. We attribute this to the superiority of LaTr's \vl backbones. As shown in \autoref{tab:comp_gitl}, LaTr\textsubscript{base} with no pre-training significantly outperforms our baseline (\baseline) on TextVQA (52.3\% vs.~45.2\%). LaTr and \ourmodel use different scene-text pre-training data: LaTr uses five times larger data than \ourmodel (64M vs.~13M in \autoref{tab:comp_sota}), which covers more \emph{diverse} scene text. This is particularly beneficial to TextVQA/ST-VQA, which contain scene text from multiple domains (\eg, brand, vehicle, etc.) and may explain why LaTr outperforms \ourmodel. 

In contrast, OCR-VQA~\cite{mishra2019ocr} only covers book-related scene text. Thus, pre-training data becomes less important than pre-training approaches, and \ourmodel outperforms LaTr (72.2\% vs.~67.5\% in \autoref{tab:comp_gitl}). Moreover, while LaTr only shows its effectiveness on VQA tasks, PreSTU shows on both VQA and image captioning tasks.

We further compare \ourmodel to extremely large-scale \vl models pre-trained on more than 2B \itform pairs. Interestingly, our best model even outperforms two much larger models Flamingo~\cite{flamingo} and GIT2~\cite{git} on some tasks; using much less data, we achieve better results than Flamingo (56.3\% vs.~54.1\%, \autoref{tab:comp_sota}) on TextVQA and than GIT2 (72.2\% vs.~69.9\%, \autoref{tab:comp_gitl}) on OCR-VQA.

Recently, PaLI~\cite{chen2022pali}, a large-scale \vl model (ViT-e/mT5-XXL) pre-trained on 10B \itform pairs, reports SOTA results on all major \vl tasks, except for VizWiz-Captions (\autoref{tab:comp_sota}). It is worth noting that 
\ourmodel (specifically, our \ourocr) was an ingredient in the pre-training objective of PaLI to tackle OCR and STU tasks, demonstrating \ourocr's utility in large-scale SOTA models.

The closest to \ourmodel in terms of model/data sizes is GIT\textsubscript{L}, a smaller-scale version of GIT2 (347M parameters and 20M \itform pairs). 
As shown in \autoref{tab:comp_gitl}, \ourmodel outperforms (or is on par with) GIT\textsubscript{L} on all tasks, demonstrating efficiency with respect to model/data sizes. See more comparisons in the supplementary material.

\begin{table*}[t]
\centering
\tabcolsep 2.5pt
\footnotesize
\begin{tabular}{ccccccccccc}
\toprule
\multirow{2}[2]{*}{Model} &
\multirow{2}[2]{*}{Vision / Language} &
\multirow{2}{*}{Model} &
\multirow{2}{*}{Data} &
\multirow{2}{*}{Pre-training} &
\multicolumn{6}{c}{Test Benchmark} \\
\cmidrule(r){6-11}
& & Size & Size & Objective & TextCaps & VizWiz-Captions & ST-VQA & TextVQA & VizWiz-VQA & VQAv2 \\
& & & & & CIDEr & CIDEr & ANLS & Acc & Acc & Acc \\
\midrule
\baseline & ViT-B/16 / mT5\textsubscript{base} & 473M & 0 & - & 96.9 & 87.2 & 56.7 & 44.8 & 57.2 & 75.2 \\
\multirow{2}{*}{{\rotatebox[origin=c]{0}{\ourmodelbf}}} 
& \multirow{2}{*}{{\rotatebox[origin=c]{0}{ViT-B/16 / mT5\textsubscript{base}}}} & \multirow{2}{*}{{\rotatebox[origin=c]{0}{473M}}} & \multirow{2}{*}{{\rotatebox[origin=c]{0}{13M}}} & \oursplitocr & 126.1 & 90.2 & 65.5 & 55.2 & 61.3 & 76.2 \\
& & & & \oursplitocrvqacap & 139.1 & 105.6 & N/A & 56.3 & 62.0 & 76.1 \\
\arrayrulecolor{gray}\hline
Non-TAP~\cite{tap} & \multirow{4}{*}{FRCNN / BERT\textsubscript{base}} & \multirow{4}{*}{146M} & 0 & - & 93.4 & - & 51.7 & 44.8 & - & - \\
TAP~\cite{tap} & & & 1.5M* & \mlmitmrpp & 109.7 & - & 59.7 & 54.0 & - & - \\
TAG~\cite{wang2022tag} & & & 88K* & \mlmitmrpp & - & - & 60.2 & 53.7 & - & - \\
LOGOS~\cite{logos} & & & 88K* & \roilocal & - & - & 57.9 & 51.1  & - & - \\
\arrayrulecolor{gray}\hline
ConCap~\cite{ConCap} & \multirow{2}{*}{BLIP} & \multirow{2}{*}{559M} & \multirow{2}{*}{129M} & \multirow{2}{*}{\lmitmitc} & 105.6 & - & - & - & - & - \\
UniTNT~\cite{UniTNT} & & & & & 109.4 & - & 66.0 & 55.4 & - & 80.1 \\
\arrayrulecolor{gray}\hline
LaTr~\cite{latr} & ViT-B/16 / T5\textsubscript{large} & 831M & 64M & \mlm & - & - & 69.6 & 61.6 & - & - \\
\arrayrulecolor{black}\midrule
Flamingo~\cite{flamingo} & NFNet / Chinchilla & 80B & 2.3B & \lm & - & - & - & 54.1 & 65.4 & 82.1 \\
GIT2~\cite{git} & DaViT / TransDec & 5B & 12.9B & \lm & 145.0 &  \textbf{120.8} & 75.8 & 67.3 & 70.1 & 81.9 \\
{\color{blue}PaLI~\cite{chen2022pali}$\dagger$} &  ViT-e / mT5-XXL & 16B & 10B & our \ourocr w/ others & \textbf{160.4} & - & \textbf{79.9} & \textbf{73.1} & \textbf{73.3} & \textbf{84.3} \\
\bottomrule
\end{tabular}
\vspace{-8pt}
\caption{\small \textbf{Comparison to prior works.}
See \S\ref{exp_sota} for a detailed discussion.
FRCNN: Faster R-CNN, TransDec: 6-layer transformer decoder, \mlm: Masked Language (visual region) Modeling, \itm: Image-Text Matching, \rpp: Relative Position Prediction, \lm: Visual Language Modeling, \itc: Image-Text Contrastive Loss, \roilocal: ROI localization. 
*: dedicated scene-text understanding data, including ST-VQA, TextVQA, TextCaps, and OCR-CC. 
\color{blue}$\dagger$: our objective \ourocr is an ingredient in their pre-training objectives.}
\vspace{-5pt}
\label{tab:comp_sota}
\end{table*}

\begin{table*}[t]
\centering
\tabcolsep 1.0pt
\footnotesize
\begin{tabular}{cccccccccccc}
\toprule
\multirow{2}[2]{*}{Model} &
\multirow{2}[2]{*}{Vision / Language} &
\multirow{2}{*}{Model} &
\multirow{2}{*}{Data} &
\multirow{2}{*}{Pre-training} &
\multicolumn{7}{c}{Val or test-dev Benchmark} \\
\cmidrule(r){6-12}
& & Size & Size & Objective & TextCaps & VizWiz-Captions & ST-VQA & TextVQA &  VizWiz-VQA & VQAv2 & OCR-VQA \\
& & & & & CIDEr & CIDEr & ANLS & Acc & Acc & Acc & Acc \\
\midrule
\baseline & ViT-B/16 / mT5\textsubscript{base} & 473M & 0 & - & 100.0 & 87.7 & 55.6 & 45.2 & 57.7 & 74.8 & 71.5 \\
\multirow{2}{*}{{\rotatebox[origin=c]{0}{\ourmodel}}} 
& \multirow{2}{*}{{\rotatebox[origin=c]{0}{ViT-B/16 / mT5\textsubscript{base}}}} & \multirow{2}{*}{{\rotatebox[origin=c]{0}{473M}}} & \multirow{2}{*}{{\rotatebox[origin=c]{0}{13M}}} & \oursplitocr & 134.6 & 90.3 & \textbf{62.7} & 55.6 & 61.9 & 76.0 & \textbf{72.2} \\
& & & & \oursplitocrvqacap & \textbf{141.7} & \textbf{105.6} & N/A & \textbf{56.7} & \textbf{62.5} & \textbf{76.1} & - \\
\midrule
LaTr\textsubscript{base}~\cite{latr} & ViT-B/16 / T5\textsubscript{base} & 281M & 0 & - & - & - & - & 52.3 & - & - & - \\
\color{Gray}{LaTr\textsubscript{base}~\cite{latr}} & \color{Gray}{ViT-B/16 / T5\textsubscript{base}} & \color{Gray}{281M} & \color{Gray}{64M} & \color{Gray}{\mlm} & \color{Gray}{-} & \color{Gray}{-} & \color{Gray}{67.5} & \color{Gray}{58.0} & \color{Gray}{-} & \color{Gray}{-} & \color{Gray}{67.5} \\
\midrule
GIT\textsubscript{L}~\cite{git} & ViT-L/14 / TransDec & 347M & 20M & \lm & 106.3 & 96.1 & 44.6 & 37.5 & \textbf{62.5} & 75.5 & 62.4 \\
\color{Gray}{GIT2~\cite{git}} & \color{Gray}{DaViT / TransDec} & \color{Gray}{5B} & \color{Gray}{12.9B} & \color{Gray}{\lm} & \color{Gray}{148.6} & \color{Gray}{119.4} & \color{Gray}{75.1} & \color{Gray}{68.4} & \color{Gray}{71.0} & \color{Gray}{81.7} & \color{Gray}{69.9} \\
\bottomrule
\end{tabular}
\vspace{-8pt}
\caption{\small \textbf{Comparison to GIT\textsubscript{L} (similar model/data sizes to \ourmodelbf).} \beer{need to modify this.} PreSTU outperforms (or is on par with) GIT\textsubscript{L} on all tasks.
{\color{gray}{GIT2}}/{\color{gray}{LaTr\textsubscript{base}-64M}} are for reference to show that PreSTU even outperforms these large-scale works on OCR-VQA.}
\vspace{-5pt}
\label{tab:comp_gitl}
\end{table*}
\subsection{Analysis}
\label{ssec:analysis}
We aim to understand \ourmodel{} in detail. We show (a) the importance of different components of our design choice, (b) its zero-shot transferability, (c) the effect of pre-training image resolution, (d) the effect of pre-training data size, and (e) the effect of downstream OCR quality.

\mypar{Detailed ablation} 
As shown in \autoref{fig:approach}, our \ourmodel{} consists of two (optional) pre-training stages, followed by fine-tuning on downstream tasks. Here, we aim to understand the gain brought by each component. We consider different combinations of the design choices at each stage and organize the results stage-by-stage into \autoref{tab:all_cases}. We have the following three major observations.

First, \oursplitocr{} is significantly and consistently better than \ourocr (Rows with \oursplitocr vs. Rows with \ourocr in their Stage-1). \ourocr is a \textbf{``pure''} OCR prediction task, a variant of our main \oursplitocr{} (OCR-conditioned OCR prediction) in which the splitting point is always at the beginning. At first glance, such a result may seem counterintuitive: predicting the entire scene text is strictly harder than predicting part of the OCR text given the other part. When thought of carefully, this result indicates that \ourocr may put too much emphasis on \emph{recognizing} scene text, at the expense of \emph{connecting} scene text to its visual context. In other words, this highlights how \oursplitocr is able to balance the two capabilities that we identify as important for STU (\S\ref{sec:intro}).

Second, \oursplitocr{} (or \ourocr{}) makes the visual component (ViT) \emph{inherently} better at recognizing text (gap between ``Yes'' and ``No'' Rows with Stage-1 pre-training vs. gap between ``Yes'' and ``No'' Rows without Stage-1 pre-training).
Without Stage-1 (\eg, \ourvqacap), removing OCR signals during fine-tuning leads to more than a 33\% drop on TextVQA and a 49 CIDEr point drop on TextCaps. 
With Stage-1, these drops become less than 17\% and 26 CIDEr points, respectively. For TextCaps, \oursplitocr with ``No'' OCR input tokens during fine-tuning even outperforms the baseline \emph{with} OCR input (116.6 vs.~100.0 in CIDEr). 
In summary, \emph{recognizing} scene text via Stage-1 pre-training is important (\ie, cannot be achieved via \ourvqa or \ourcap alone).

Third, having two sources of OCR signals is beneficial. OCR signals by pre-trained ViT (Row \oursplitocrvqacap{} with ``No'') and OCR signals by the off-the-shelf system (Row \baseline "Yes") are complementary; we achieve the best result when leveraging both OCR signal sources (Row \oursplitocrvqacap{} with ``Yes''). See more ablation studies in the supplementary material.

\mypar{Zero-shot transferability on scene-text VQA}
\autoref{tab:zero_shot} shows zero-shot transferability of \oursplitocr{} on TextVQA. We observe that performing \oursplitocr{} and then fine-tuning on ST-VQA (\oursplitocrvqa) already leads to a strong model; \oursplitocrvqa \emph{without} fine-tuning (44.3\%) is competitive to \baseline \emph{with} fine-tuning on TextVQA training set (45.2\%), while ST-VQA alone (\ourvqa{}) only achieves 35.7\%. This suggests that \oursplitocr{} enables generalization for STU and may remove the need to collect TextVQA data entirely!

\begin{table}[t]
\centering
\small
\tabcolsep 3.5pt
\renewcommand\arraystretch{1.0}
\begin{tabular}{ccccc}
\toprule
\multicolumn{2}{c}{Pre-training} & Fine-tuning & TextVQA & TextCaps \\
Stage-1 & Stage-2 & OCR input & Val Acc & Val CIDEr \\
\midrule
\multirow{2}{*}{{{\rotatebox[origin=c]{0}{-}}}} & \multirow{2}{*}{{{\rotatebox[origin=c]{0}{-}}}} & No & 19.5 & 40.1 \\
& & Yes & 45.2 & 100.0 \\
\midrule
\multirow{2}{*}{{{\rotatebox[origin=c]{0}{-}}}} & \multirow{2}{*}{{{\rotatebox[origin=c]{0}{\ourvqacap}}}} & No & 13.7 & 81.1 \\
& & Yes & 47.2 & 130.2 \\
\midrule
\multirow{2}{*}{{{\rotatebox[origin=c]{0}{\ourocr}}}} & \multirow{2}{*}{{{\rotatebox[origin=c]{0}{-}}}} & No & 35.8 & 110.4 \\
& & Yes & 49.9 & 126.7 \\
\midrule
\multirow{2}{*}{{{\rotatebox[origin=c]{0}{\ourocr}}}} & \multirow{2}{*}{{{\rotatebox[origin=c]{0}{\ourvqacap}}}} & No & 38.6 & 108.9 \\
& & Yes & 51.9 & 134.4 \\
\midrule
\multirow{2}{*}{{{\rotatebox[origin=c]{0}{\oursplitocr}}}} & \multirow{2}{*}{{{\rotatebox[origin=c]{0}{-}}}}& No & 39.4 & 116.6 \\
& & Yes & 55.6 & 134.6 \\
\midrule
\multirow{2}{*}{{{\rotatebox[origin=c]{0}{\oursplitocr}}}} & \multirow{2}{*}{{{\rotatebox[origin=c]{0}{\ourvqacap}}}} & No & 44.3 & 118.4 \\
& & Yes & \textbf{56.7} & \textbf{141.7} \\
\bottomrule
\end{tabular}
\vspace{-5pt}
\caption{\small \textbf{Main ablation studies} for validating the importance of our main components: \oursplitocr{}, \ourvqacap{}, and having OCR input during fine-tuning. See \S\ref{ssec:analysis} for a detailed discussion. \ourocr refers to predicting the entire OCR text.}
\vspace{-5pt}
\label{tab:all_cases}
\end{table}

\mypar{Effect of image resolutions during pre-training}
We hypothesize that pre-training with high-resolution images is important for scene-text recognition; \autoref{tab:img_res} supports this argument. Further, pre-training with the 224x224 image resolution (standard resolution for many vision tasks) almost does not help;
it achieves the accuracy of 47.1\%, close to 45.2\% of \baseline baseline (\autoref{tab:all_cases} Row 2), suggesting non-standard resolution must be considered to reap the benefit of STU pre-training. 

\mypar{Effect of pre-training data scale}
How much data do we need to learn to recognize text?
\autoref{tab:data_scale} shows the performance of TextVQA given checkpoints pre-trained on 1\%, 3\%, 10\%, and 30\% subsets of CC15M. We find that the TextVQA performance goes up as more pre-training data is included. This highlights the importance of data scale in acquiring \emph{transferable} scene-text recognition skills.

\begin{table}[t]
\centering
\small
\tabcolsep 3.5pt
\renewcommand\arraystretch{1.0}
\begin{tabular}{cccc}
\toprule
\multirow{2}{*}{Model} & Pre-training & \multirow{2}{*}{Fine-tuning} & TextVQA \\
& Objective & & Val Acc \\
\midrule
\multirow{2}{*}{\baseline} & \multirow{2}{*}{-} & - & 0.04 \\
& & TextVQA & 45.2 \\
\midrule
\multirow{2}{*}{\ourmodel} & \ourvqa & - & 35.7 \\
& \oursplitocrvqa & - & 44.3 \\

\bottomrule
\end{tabular}
\vspace{-5pt}
\caption{\small \textbf{Zero-shot transferability on TextVQA.} Our zero-shot \oursplitocrvqa (\emph{without} fine-tuning on TextVQA) is competitive to supervised \baseline (\emph{with} fine-tuning on TextVQA).}
\label{tab:zero_shot}
\end{table}

\begin{table}[t]
\centering
\small
\tabcolsep 3.5pt
\renewcommand\arraystretch{0.8}
\begin{tabular}{ccccc}
\toprule
\multirow{2}[2]{*}{Model} &
\multicolumn{2}{c}{Pre-training} &
\multirow{2}{*}{Fine-tuning} &
\multirow{2}{*}{TextVQA} \\
\cmidrule(r){2-3}
& Objective & Resolution & Resolution & Val Acc\\
\midrule
\multirow{4}{*}{{\rotatebox[origin=c]{0}{\ourmodel}}}
& \multirow{4}{*}{{\rotatebox[origin=c]{0}{\oursplitocr}}} & 224 & \multirow{4}{*}{{\rotatebox[origin=c]{0}{640}}} & 47.1 \\
& & 384 & & 50.2 \\
& & 480 & & 53.1 \\
& & 640 & & \textbf{55.6} \\
\bottomrule
\end{tabular}
\vspace{-5pt}
\caption{\small \textbf{Effects of image resolutions.} TextVQA accuracy goes up as the pre-training image resolution increases, emphasizing the necessity
 of high-resolution images during pre-training.}
\vspace{-5pt}
\label{tab:img_res}
\end{table}

\begin{table}[t]
\centering
\small
\tabcolsep 3.5pt
\renewcommand\arraystretch{1.0}
\begin{tabular}{ccccc}
\toprule
\multirow{2}[2]{*}{Model} &
\multicolumn{3}{c}{Pre-training} &
\multirow{2}{*}{TextVQA} \\
\cmidrule(r){2-4}
& Objective & Proportion & \# of Data & Val Acc \\
\midrule
\multirow{5}{*}{{\rotatebox[origin=c]{0}{\ourmodel}}}
& \multirow{5}{*}{{\rotatebox[origin=c]{0}{\oursplitocr}}} & 1\% & 130K & 42.3 \\
& & 3\% & 390K & 45.4 \\
& & 10\% & 1.3M & 50.6 \\
& & 30\% & 3.9M & 53.0 \\
& & 100\% & 13M & \textbf{55.6} \\
\bottomrule
\end{tabular}
\vspace{-5pt}
\caption{\small \textbf{Importance of pre-training data scale.}
TextVQA performance improves as more pre-training data, showing the importance of data scale in learning \emph{transferable} scene-text recognition.}
\vspace{-5pt}
\label{tab:data_scale}
\end{table}

\mypar{Effect of downstream OCR systems} 
We study the effect of different OCR systems during fine-tuning (\autoref{tab:ocr_sys}). We observe that the \oursplitocr-pre-trained model is more robust to the change in downstream OCR systems than \baseline. Indeed, \oursplitocr + Rosetta can even perform better than \baseline + gOCR. This result is consistent with \autoref{tab:all_cases}, where we experiment with removing OCR texts entirely during fine-tuning.
We also find that gOCR is the most effective. Interestingly, it is even better than human-annotated TextOCR; we hypothesize this is because TextOCR only provides word-level annotation whereas gOCR provides some grouping.

\begin{table}[t]
\centering
\small
\tabcolsep 4pt
\renewcommand\arraystretch{1.0}
\begin{tabular}{cccc}
\toprule
\multirow{2}{*}{Model} & Pre-training & Fine-tuning & TextVQA \\
& Objective & OCR System & Val Acc \\
\midrule
\multirow{3}[1]{*}{{\rotatebox[origin=c]{0}{\baseline}}}
& \multirow{3}{*}{{{\rotatebox[origin=c]{0}{-}}}} & TextOCR~\cite{textcaps} & 44.0 \\
& & Rosetta~\cite{borisyuk2018rosetta} &  36.7 \\ 
& & gOCR & 45.2 \\
\midrule
\multirow{3}[1]{*}{{\rotatebox[origin=c]{0}{\ourmodel}}} & \multirow{3}{*}{{{\rotatebox[origin=c]{0}{\oursplitocr}}}} & TextOCR~\cite{textcaps} & 54.8 \\
& & Rosetta~\cite{borisyuk2018rosetta} &  50.7 \\ 
& & gOCR & \textbf{55.6} \\
\bottomrule
\end{tabular}
\vspace{-5pt}
\caption{\small \textbf{Effect of downstream OCR systems on TextVQA.} \oursplitocr makes the model more robust to the change in OCR systems during fine-tuning.}
\vspace{-15pt}
\label{tab:ocr_sys}
\end{table}

\section{Related Work}
\label{sec:related}

\mypartop{Scene-Text Understanding}
Most early STU works~\cite{jaderberg2014synthetic,jaderberg2016reading,li2017towards,borisyuk2018rosetta,he2018end,liu2018fots} have merely focused on Optical Character Recognition (OCR). We instead focus on scene-text understanding (STU) in the context of \vl{} tasks: VQA~\cite{textvqa,stvqa} and image captioning~\cite{textcaps}. The most common approach for these STU tasks is to fuse pre-extracted object detection features with off-the-shelf OCR signals as additional input~\cite{textvqa,m4c,textcaps,latr,local-aware,spatially-aware,wang2022tag,xu2023device,SceneGATE,li2022two}. 
These works often focus on specific challenges in downstream STU tasks, including dealing with noisy OCR signals, enabling the generation of rare words, or incorporating geometric information of OCR texts.
In contrast, our work focuses on pre-training general-purpose STU models and shows the effectiveness of our objectives on multiple downstream STU tasks (\S\ref{ssec:exp_res}).

\mypar{\vl Pre-Training for STU}
One line of works incorporates OCR signals explicitly for pre-training~\cite{tap, latr, logos}. TAP proposes an objective to learn the relative spatial position of two OCR texts.
LOGOS~\cite{logos} localizes a region that is most related to a given task and relies on its OCR text to complete the task.
LaTr~\cite{latr} models the co-occurrence statistics of layout-aware OCR tokens. Our pre-training objectives, on the other hand, focus on learning both scene-text recognition and the role of scene-text in its visual context.

The other line of works is OCR-free.
Recently, extremely large image-text models have shown promising results on STU tasks, despite having no explicit STU objectives
(\eg, GIT2~\cite{git}, Flamingo~\cite{flamingo}). 
However, it would require an analysis of their private data and a prohibitive amount of resources to pinpoint what contributes to such strong results. Our study offers a complementary perspective to this OCR-free approach by pushing the limit of the OCR-heavy approach further than before and conducting more thorough experiments at a smaller scale.
\section{Conclusion}
\label{sec:conclusion}

We introduce a simple recipe for scene-text understanding, consisting of  OCR-aware pre-training objectives operating from image pixels. Our task-agnostic objective \oursplitocr teaches the model to recognize scene text and to connect scene text to its visual context. Our task-specific objectives \ourvqa and \ourcap further strengthen that connection. We conduct comprehensive experiments to demonstrate the utility of this recipe.

{\normalsize{
\mypar{Acknowledgments}
We would like to thank Bo Pang, Xiao Wang, Kenton Lee, and Tania Bedrax-Weiss for their thoughtful feedback and discussions. J. Kil and W. Chao are supported in part by grants from the National Science Foundation (IIS-2107077, OAC-2118240, and OAC-2112606) and Cisco Systems, Inc.}}

{\small
\bibliographystyle{ieee_fullname}
\bibliography{main}
}

\newpage
\appendix
\appendix
\section*{Appendices}
In this supplementary material, we provide details omitted in the main text.
\begin{itemize}
    [itemsep=1.5pt,topsep=1.5pt,leftmargin=5pt]
    \item \autoref{apdx:imp_details}: \vl model implementation details (cf.~\S\ref{setup} of the main text).
    
    \item \autoref{apdx:dataset}: Pre-training \& Scene-text \vl datasets (cf.~\S\ref{data} \&~\S\ref{finetune_stage} of the main text).
    
    \item \autoref{apdx:more_comp_prior_works}: More comparisons to prior works (cf.~\S\ref{exp_sota} of the main text).
    
    \item \autoref{apdx:more_ablation}: More ablation studies (cf.~\S\ref{ssec:analysis} of the main text).
    
    \item \autoref{apdx:more_qual_res}: Qualitative results.

    \item \autoref{apdx:contribute}: Contributions.
    
\end{itemize}

\section{\vl model implementation details}
\label{apdx:imp_details}
Our model is an encoder-decoder \vl architecture consisting of ViT-B/16~\cite{vit} as a visual module and mT5-Base~\cite{mt5} as a language module.
For the vision module, we adopt a transformer-based vision model ViT~\cite{vit} pre-trained on JFT-3B dataset~\cite{jft-3B}, the extension of JFT-300M~\cite{jft300m}, with 3 billion images collected from the web. 
Our language module is initialized from mT5-Base~\cite{mt5}, a multilingual variant of T5~\cite{t5}, pre-trained on a new Common Crawl-based dataset with 101 different languages. 

During training, all parameters in vision and language blocks are updated simultaneously. We choose Adafactor~\cite{adafactor} as an optimizer with $\beta_1$ = 0 and second-moment exponential decay = 0.8. For a learning rate, we schedule a linear warmup for 1K steps with inverse square-root decay. Our \vl architecture is implemented in Jax/Flax~\cite{jax/flax} based on the open-source T5X~\cite{t5x} framework. 

We have done extensive hyperparameter tuning for our experiments. For instance, we find that the best hyper-parameter configuration for \oursplitocr pre-training is --- initial (peak) learning rate: 1e-3, batch size: 256, image resolution: 640x640, the length of input/target text tokens: 40/26, and dropout: 0.1. For TextVQA, we achieve the best result with initial learning rate: 2e-4 and the length of input/target text tokens: 72/8 (See \autoref{apdx:hyper_param} for more details).

\begin{table*}[t]
\centering
\small
\tabcolsep 4.5pt
\renewcommand\arraystretch{1.0}
\begin{tabular}{cccccccc}
\toprule
\multirow{2}[1]{*}{Hyper-parameter} &
\multicolumn{1}{c}{Pre-training} &
\multicolumn{6}{c}{Downstream} \\
\cmidrule(r){2-2}
\cmidrule(r){3-8}
& \oursplitocr & ST-VQA & TextVQA & VizWiz-VQA & VQAv2 & TextCaps & VizWiz-Caption \\
\midrule
Initial (peak) learning rate & 1e-3 & 9e-4 & 2e-4 & 9e-4 & 1e-3 & 2e-4 & 2e-4 \\
Batch size & 256 & 256 & 256 & 256 & 512 & 256 & 256 \\
Image resolution & 640x640 & 640x640 & 640x640 & 640x640 & 640x640 & 640x640 & 640x640 \\
Length of input text tokens & 40 & 72 & 72 & 72 & 72 & 56 & 56 \\
Length of target text tokens & 26 & 8 & 8 & 8 & 8 & 64 & 64 \\
Dropout & 0.1 & 0.1 & 0.1 & 0.1 & 0.1 & 0.1 & 0.1 \\
\bottomrule
\end{tabular}
\captionsetup{width=.95\textwidth}
\vspace{-5pt}
\caption{\small \textbf{Best hyper-parameters for our experiments.} Among hyper-parameters of our \vl model, we find that initial (peak) learning rate, batch size, image resolution, length of input/target text tokens, and dropout are major components affecting the performance of our tasks.}
\label{apdx:hyper_param}
\end{table*}

\begin{table*}[t]
\centering
\tabcolsep 1.5pt
\footnotesize
\begin{tabular}{ccccccccccc}
\toprule
\multirow{2}[2]{*}{Model} &
\multirow{2}[2]{*}{Vision / Language} &
\multirow{2}{*}{Model} &
\multirow{2}{*}{Data} &
\multirow{2}{*}{Pre-training} &
\multicolumn{6}{c}{Scene-text \vl Benchmark} \\
\cmidrule(r){6-11}
& & Size & Size & Objective & ST-VQA & TextVQA & VizWiz-VQA & VQAv2 & TextCaps & VizWiz-Captions \\
& & & & & ANLS & Acc & Acc & Acc & CIDEr & CIDEr \\
\midrule
\baseline
& ViT-B16 / mT5\textsubscript{base} & 473M & 0 & - & 56.7 (55.6) & 44.8 (45.2) & 57.2 (57.7) & 75.2 (74.8) & 96.9 (100.0) & 87.2 (87.7) \\
\midrule
\multirow{3}{*}{{\rotatebox[origin=c]{0}{\ourmodel}}}
& \multirow{3}{*}{{\rotatebox[origin=c]{0}{ViT-B16 / mT5\textsubscript{base}}}} & \multirow{3}{*}{{\rotatebox[origin=c]{0}{473M}}} & \multirow{3}{*}{{\rotatebox[origin=c]{0}{13M}}} & \ourvqacap & N/A (N/A) & 48.3 (47.2) & 57.6 (58.3) & 75.0 (75.0) & 133.1 (130.2) & 103.1 (103.6) \\
& & & & \oursplitocr & 65.5 (62.7) & 55.2 (55.6) & 61.3 (61.9) & 76.2 (76.0) &126.1 (134.6) & 90.2 (90.3) \\
& & & & \oursplitocrvqacap & N/A (N/A) & 56.3 (56.7) & 62.0 (62.5) & 76.1 (76.1) & 139.1 (141.7) & 105.6 (105.6) \\
\midrule
TAP~\cite{tap} & FRCNN / BERT\textsubscript{base} & 146M & 1.5M & \mlmitmrpp & 59.7 (59.8) & 54.0 (54.7) & - (-) & - (-) & 109.7 (119.0) & - (-) \\
LaTr~\cite{latr} & ViT-B/16 / T5\textsubscript{large} & 831M & 64M & \mlm & 69.6 (70.2) & 61.6 (61.1) & - (-) & - (-) & - (-) & - (-) \\
\midrule
Flamingo~\cite{flamingo} & NFNet / Chinchilla & 80B & 2.3B & \lm & - (-) & 54.1 (57.1) & 65.4 (65.7) & 82.1 (82.0) & - (-) & - (-) \\
GIT\textsubscript{L}~\cite{git} & CoSwin / TransDec & 347M & 20M & \lm & - (44.6) & - (37.5) & - (62.5) & - (75.5) & - (106.3) & - (96.1)\\
GIT2~\cite{git} & DaViT / TransDec & 5B & 12.9B & \lm & 75.8 (75.1) & 67.3 (68.4) & 70.1 (71.0) & 81.9 (81.7) & 145.0 (148.6) & 120.8 (119.4) \\
\color{blue}PaLI~\cite{chen2022pali}$\dagger$ & ViT-e / mT5-XXL & 16B & 10B & our \ourocr w/ others  & 79.9 (-) & 73.1 (71.8) & 73.3 (74.4) & 84.3 (84.3) & 160.4 (160.0) & - (-) \\

\bottomrule
\end{tabular}
\vspace{-5pt}
\caption{\textbf{Full Comparison to prior works.} FRCNN: Faster R-CNN, TransDec: 6-layer transformer decoder, \mlm: Masked Language (visual region) Modeling, \itm: Image-Text Matching, \rpp: Relative Position Prediction, \lm: Visual Language Modeling. 
Following~\cite{git}, the parameters of text token embeddings are not counted in the model size. 
We report results on the \textbf{test (validation)} set for ST-VQA, the \textbf{test-std (validation)} for TextVQA/TextCaps, and the \textbf{test-std (test-dev)} set for VizWiz-VQA, VQAv2, and VizWiz-Captions. \color{blue}$\dagger$: our objective \ourocr is an ingredient in their pre-training objectives.}
\vspace{-10pt}
\label{apdx:comp_sota_test_val}
\end{table*}

\section{Pre-training \& Scene-text \vl datasets}
\label{apdx:dataset}
We provide more details about pre-training and scene-text \vl datasets used in our experiments.

\textbf{Scene-Text on CC15M}. We estimate the portion of scene text on CC15M with a study on $300$ randomly sampled images. We manually check each image and found: $59\%$ (177/300) have scene text; only 13\% (38/300) are watermark-only images. This aligns with TAP's report~\cite{tap} on CC3M (scene-text: 42\%, watermark-only: 5\%). Note that TAP mentioned \emph{``only the CC dataset contains a reasonable portion of images with meaningful scene text regions''}, suggesting CC15M is suitable for STU pre-training.

\textbf{ST-VQA}~\cite{stvqa} is for scene-text VQA dataset. Its images are collected from various resources: COCO-Text~\cite{veit2016coco}, Visual Genome~\cite{krishna2017visual}, VizWiz~\cite{vizwiz}, ICDAR~\cite{karatzas2013icdar,karatzas2015icdar}, ImageNet~\cite{deng2009imagenet}, and IIIT-STR~\cite{mishra2013image}. Since there is no official validation set, we follow the split provided by M4C~\cite{m4c}, resulting in 23K/26K training/validation VQA examples.

\textbf{TextVQA}~\cite{textvqa} for scene-text VQA. It is a subset of Open Images~\cite{openimages} with scene-text related QA pairs from human annotators with ten ground-truth answers. It has 34K/5K training/validation VQA examples from 21K/3K images.

\textbf{VizWiz-VQA}~\cite{vizwiz}. The dataset contains 20K/3K training/validation VQA examples collected from blind users. Due to the nature of the questions asked by blind people, we identify this benchmark as a candidate to benefit from scene-text understanding, even though it was not directly designed for scene-text VQA.

\textbf{VQAv2}~\cite{vqav2}. We further evaluate \ourmodel on standard VQA benchmark to check if the scene-text recognition can also help on general VQA tasks. Following~\cite{jiang2018pythia}, we use the VQAv2 train/dev splits of *train2014/minival2014, which are 592K/65K VQA examples in total.  

\textbf{TextCaps}~\cite{textcaps} for scene-text image captioning task. It uses the same subset of OpenImages images with TextVQA. Each image has five ground-truth captions, totaling 100K/15K training/validation captions.

\textbf{VizWiz-Captions}~\cite{vizwizcap}. Like Vizwiz-VQA, this benchmark was generated by blind users to solve their daily visual challenges. It contains 23.4K/7.7K training/validation images, where each image is paired with five captions. In total, there are 117K/38K training/validation image captions.

\textbf{OCR-VQA}~\cite{mishra2019ocr} is an OCR-based VQA dataset about images of book covers. Concretely, it requires models to answer visual questions by reading/interpreting the text on the book covers (\eg, author, title). In summary, OCR-VQA provides 207K images of book covers and more than 1 million VQA examples. 

\textbf{DocVQA}~\cite{docvqa} asks for the textual (handwritten, typewritten, printed) content on the document images. In contrast with general VQA~\cite{vqav2}, models should understand additional visual cues, including layout (\eg, tables), style (\eg, font, color), and non-textual elements (\eg, tick boxes). In total, DocVQA contains 50K VQA examples with more than 12K document images.

\textbf{ChartQA}~\cite{chartqa} is a VQA benchmark based on charts. Specifically, it covers more than 23K VQA examples from 17K charts. In ChartQA, models are required to perform complex reasoning (\eg, logical and arithmetic operations) to understand charts and the corresponding questions.

\textbf{AI2D}~\cite{AI2D} is a VQA dataset of illustrative diagrams. The task of AI2D is to answer diagram-related questions by analyzing the diagram structure and identifying its visual entities and their semantic relationships. AI2D provides 5K diagrams with 15K VQA examples in total.

\textbf{WidgetCap}~\cite{widgetcaptioning} aims to generate language descriptions for UI elements (widgets) in the mobile interface. Mobile apps often lack widget captions in their interfaces, which recently becomes a primary issue for mobile accessibility. WidgetCap attempts to solve this challenge by providing an evaluation benchmark containing more than 162K language phrases (\ie, captions) with 61K UI elements.

\textbf{Screen2Words}~\cite{screen2words} is an image captioning task to generate a short summary of the mobile screen. To complete the task, models should have the capability of understanding the screen and conveying its content and functionalities in a concise language phrase. Screen2Words consists of 112K captions for 22K mobile screens in total.

\section{More comparisons to prior works}
\label{apdx:more_comp_prior_works}
\textbf{Comparison to TAP}. While \ourmodel adopts a \emph{general} pre-training dataset (\ie, CC15M), TAP's pre-training data aggregates scene-text \emph{dedicated} downstream data, including ST-VQA, TextVQA, TextCaps, and OCR-CC. Thus, even if the size of TAP's pre-training data (1.5M) is smaller, it may align better with the downstream tasks. However, since TAP's approach focuses on the specific downstream tasks, it is less applicable to other \vl tasks, whereas \ourmodel provides a more flexible interface.

Moreover, TAP adopts closed-set prediction by training an answer classifier based on the dataset-specific vocabulary. This may benefit the accuracy of the corresponding downstream task. In contrast, \ourmodel chooses open-ended prediction as it is more generalizable in practice and is adopted by many recent works (\eg, PaLI, GIT). 

\textbf{Full Comparison}. \autoref{apdx:comp_sota_test_val} shows full comparisons to prior works on all splits of benchmarks. Concretely, we report results on the \textbf{test (validation)} set for ST-VQA,
the \textbf{test-std (validation)} for TextVQA/TextCaps, and the \textbf{test-std (test-dev)} set for VizWiz-VQA, VQAv2, and VizWiz-Captions. Aligned with the results in the main text, \oursplitocr outperforms \baseline on all evaluation metrics. In addition, \oursplitocrvqacap further boosts the performance, highlighting the importance of task-specific objectives (\ourvqa and \ourcap) during pre-training.

\begin{figure}
    \centering
    \centerline{\includegraphics[width=\linewidth]{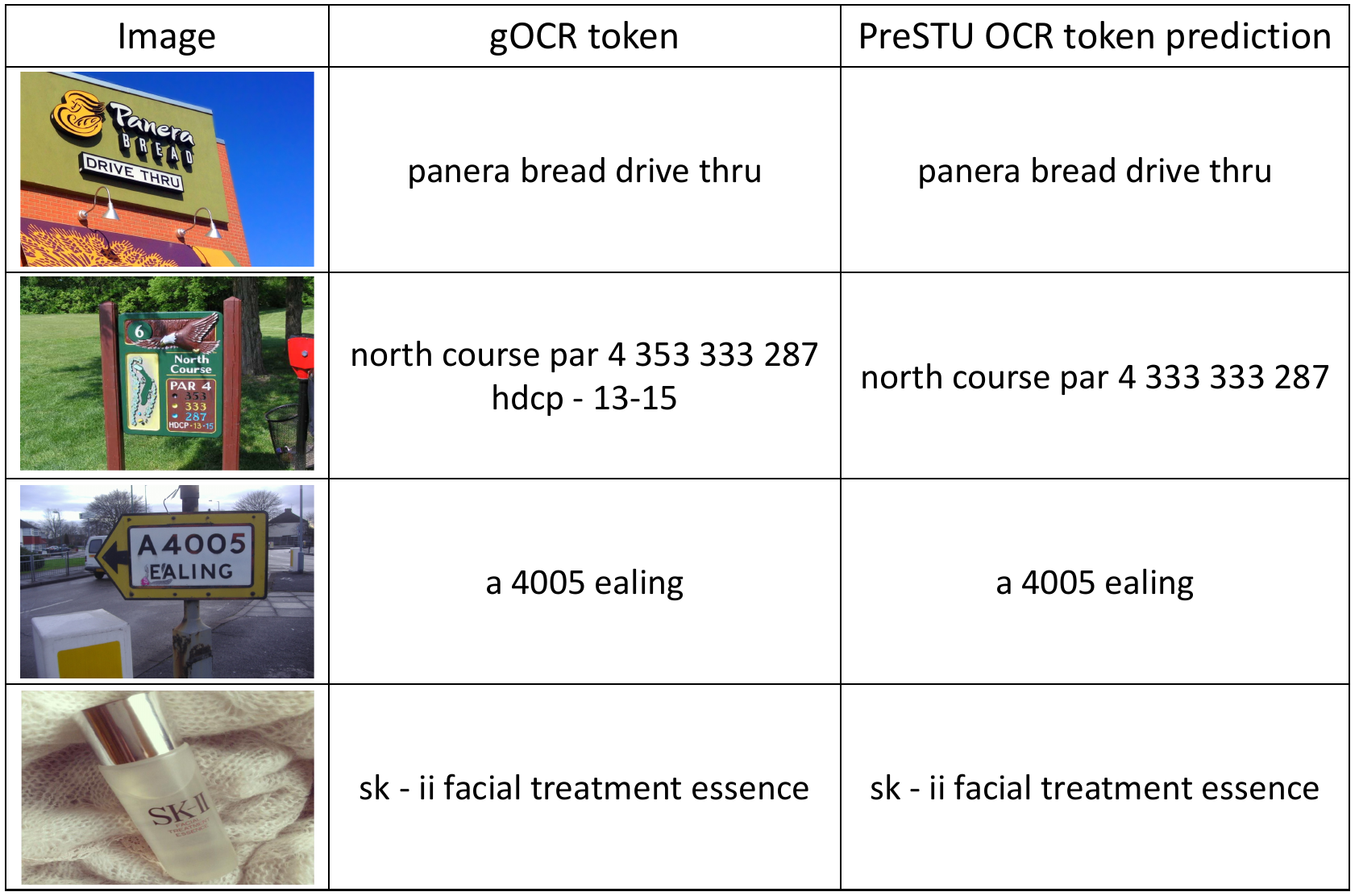}}
    \vspace{-5pt}
    \caption{\textbf{\ourmodelbf's OCR token prediction.} 
    The quality of OCR tokens generated by \oursplitocr is comparable to that of gOCR system. 
    This shows the possibility of leveraging \oursplitocr as an alternative OCR system when other systems are not available.}
    \label{apdx:ocr_pred}
\end{figure}

\begin{figure}
    \centering
    \centerline{\includegraphics[width=\linewidth]{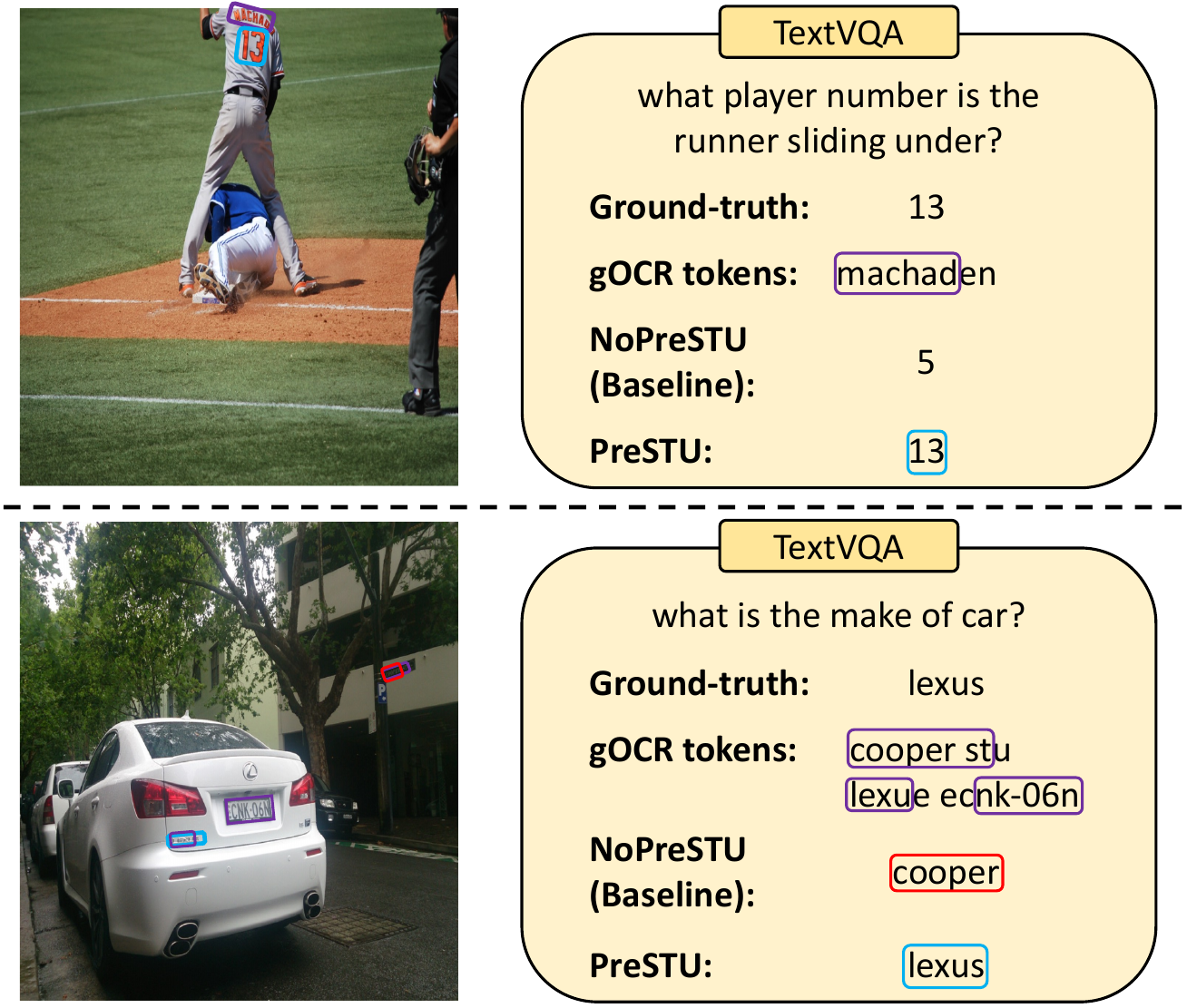}}
    \vspace{-5pt}
    \caption{\textbf{gOCR tokens vs. \ourmodelbf prediction on TextVQA.} gOCR system does not detect some OCR tokens in the image (\eg, ``13'') or detects them incorrectly (\eg, ``lexue''). This leads \baseline to predict wrong answers (\eg, ``5'' or ``cooper''). On the other hand, \oursplitocr with gOCR tokens as input predicts the answers correctly with correct OCR tokens (\eg, ``13'' or ``lexus'').}
    \label{apdx:gocr}
    \vspace{-10pt}
\end{figure}

\section{More ablation studies}
\label{apdx:more_ablation}
\textbf{\oursplitocrbf vs.~\ourcapbf}. \autoref{tab:main_vqa} of the main text shows the effectiveness of \oursplitocr against \ourvqa on VQA tasks. We further check its benefit over \ourcap on VQA tasks. As shown in \autoref{apdx:splitocr_vs_cap}, \oursplitocr consistently improves over \ourcap (\eg, 53.2\% vs.~49.3\%) on TextVQA, further supporting that \oursplitocr is important for higher accuracy.

We also investigate the effect of the order of pre-training stages. Concretely, we switch the order between \oursplitocr and \ourcap and demonstrate that applying \oursplitocr first (\ie, default setting) is better (\autoref{apdx:switch_stage}).

\begin{table}[t]
\centering
\small
\tabcolsep 5.5pt
\renewcommand\arraystretch{1.0}
\begin{tabular}{ccc}
\toprule
\multirow{2}{*}{Model} &
\multirow{2}[-2]{*}{Pre-training} & \multirow{2}[-2]{*}{TextVQA} \\
& Objective & Val Acc \\
\midrule
\multirow{4}{*}{{\rotatebox[origin=c]{0}{\ourmodel}}}
& \ourcap & 49.3 \\
& \oursplitocrcap & 53.2 \\
& \ourcapvqa & 50.0 \\
& \oursplitocrcapvqa & 55.0 \\
\bottomrule
\end{tabular}
\vspace{-5pt}
\caption{\small \textbf{\oursplitocrbf vs.~\ourcapbf on VQA tasks.} \oursplitocr is crucial for higher accuracy.}
\label{apdx:splitocr_vs_cap}
\end{table}

\begin{table}[t]
\centering
\small
\tabcolsep 5.5pt
\renewcommand\arraystretch{1.0}
\begin{tabular}{ccc}
\toprule
\multirow{2}{*}{Model} &
\multirow{2}[-3]{*}{Pre-training} & \multirow{2}[-3]{*}{TextCaps} \\
& Objective & Val CIDEr \\
\midrule
\multirow{2}{*}{{\rotatebox[origin=c]{0}{\ourmodel}}}
& \oursplitocrcap & 141.7 \\
& \ourcapsplitocr & 135.4 \\
\bottomrule
\end{tabular}
\vspace{-5pt} 
\caption{\small \textbf{Effect of switching pre-training stages.} Applying \oursplitocr first (\ie, default setting) is more effective.}
\vspace{-5pt}
\label{apdx:switch_stage}
\end{table}

\textbf{Order of OCR}. \ourmodel uses the fixed OCR order to standardize the target output sequence during pre-training. Compared to the random order, we see its advantage with consistent improvements (\eg, 132.4 vs.~134.6 on TextCaps CIDEr / 55.3\% vs.~55.6\% on TextVQA).

\textbf{OCR System}. We note that different prior works often use different \emph{commercial} OCR engines to obtain their best results. Thus, it is hard to perform a fair comparison without extra costs. 
That said, we did evaluate \ourmodel with different OCR engines (including Rosetta-en) at the downstream stage (\autoref{tab:ocr_sys} of the main text). 
A similar setup is used in LaTr~\cite{latr}: Rosetta-en/Amazon-OCR for downstream TextVQA/pre-training, respectively. In this setup, \ourmodel outperforms LaTr on TextVQA Val (50.7\% vs.~48.4\%).

\section{Qualitative results}
\label{apdx:more_qual_res}

\autoref{apdx:ocr_pred} shows some examples of OCR tokens generated by \oursplitocr. Our \oursplitocr detects all (or almost all) OCR tokens in the images correctly, competitive to the gOCR system.

In~\S\ref{ssec:analysis} of the main text, we demonstrate that having two sources of OCR signals is beneficial (OCR signals by pre-trained ViT with \oursplitocr and OCR signals by gOCR system). \autoref{apdx:gocr} further supports this finding qualitatively. For instance, gOCR alone does not detect some OCR tokens in the image (\eg, ``13'') or detects them incorrectly (\eg, ``lexue''). This leads \baseline to predict wrong answers (\eg, ``5'' or ``cooper''). On the other hand, \oursplitocr with gOCR tokens as input predicts the answers correctly with correct OCR tokens (\eg, ``13'' or ``lexus''), demonstrating that two sources of OCR signals (\ie, ViT and gOCR) are complementary.

\begin{figure}
    \centering
    \centerline{\includegraphics[width=\linewidth]{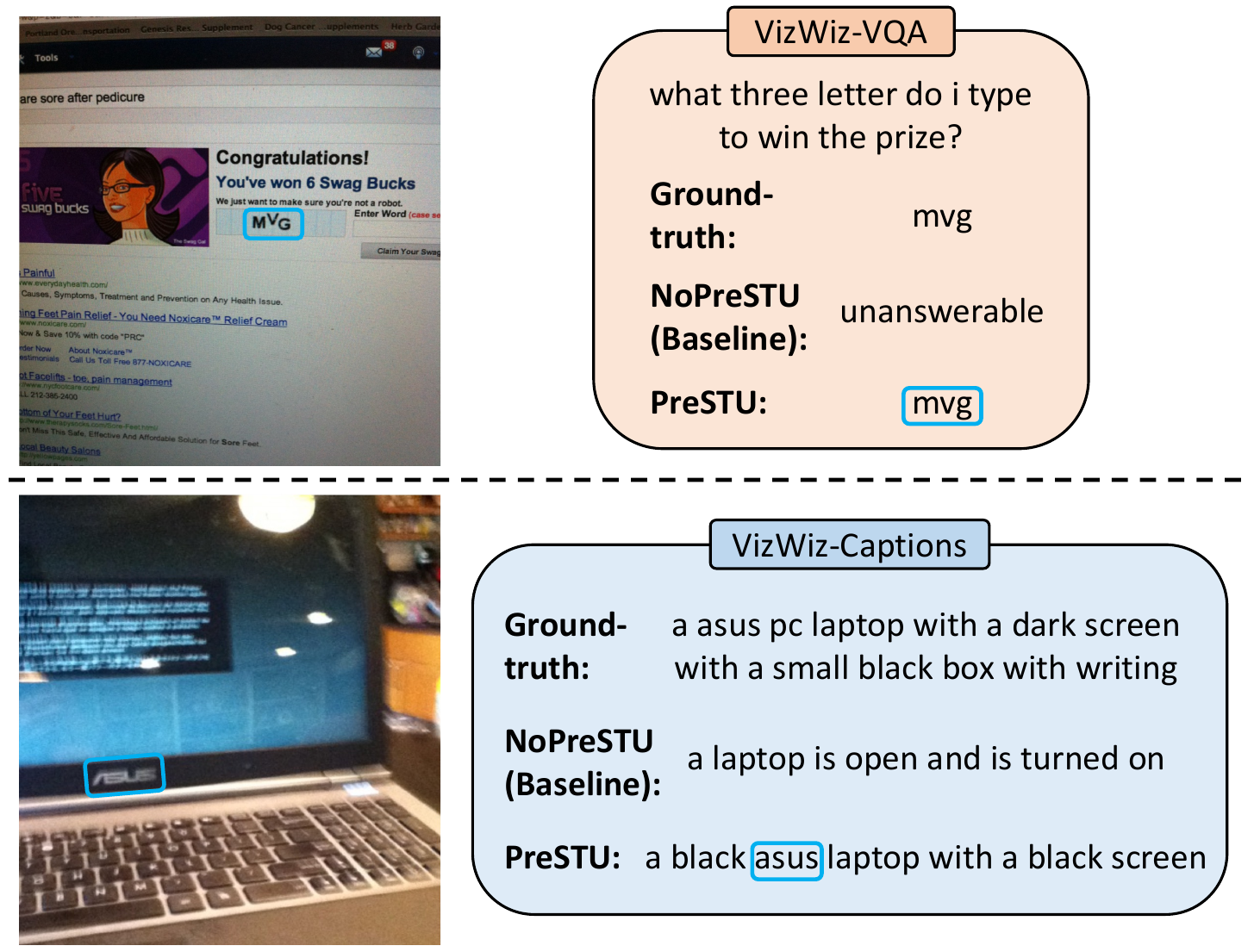}}
    \vspace{-5pt}
    \caption{\small \textbf{Qualitative results on VizWiz-VQA~\cite{vizwiz} and VizWiz-Captions~\cite{vizwizcap}.}}
    \label{fig:qual_vizwiz}
\vspace{-10pt}
\end{figure}

\autoref{fig:qual_vizwiz} provides qualitative results for VizWiz-VQA and VizWiz-Captions, demonstrating the applicability of \ourmodel to different VQA and image captioning tasks.

\section{Contributions}
\label{apdx:contribute}
While our \oursplitocr is inspired by SimVLM~\cite{wang2022simvlm}, the motivation is fundamentally different and it is not trivial to apply the prefix idea in the first place for OCR-aware pre-training. Concretely, SimVLM aims to serve downstream tasks that generate text like captions or answers (with optional text input). Thus, it is understandable why SimVLM could help. In contrast, for downstream STU tasks, \emph{OCR strings often serve only as the text input (Figures 2 \& 3 of the main text).}
Therefore, while it makes sense to apply our second stage pre-training (\ourcap \& \ourvqa) with OCR strings as the input, it is not intuitive to develop a separate OCR-only pre-training stage (\oursplitocr) that leverages the idea of SimVLM. We came up with \oursplitocr purely from the two essential STU capabilities: (i) recognizing text in an image, (ii) connecting the text to its visual context.
Our contribution thus lies in how to fulfill the two requirements via a unified manner, which turns out to be a SimVLM-like objective.

Besides \oursplitocr, another key contribution of our work is the comprehensive investigation of pre-training STU capabilities using a combination of easily reproducible objectives and a standard network architecture, on domains much more diverse than in previous works.
Thus, we believe that our extensive analysis is valuable to the community.

Finally, we demonstrate the effectiveness of our OCR-aware method in large-scale settings. We choose CC15M as the pre-training dataset, which is often considered large-scale, and PaLI~\cite{chen2022pali}, an extremely large-scale model (with $10$B data), utilizes our objective to achieve SOTA results on nearly all STU tasks (cf.~\S\ref{exp_sota} of the main text). This shows the utility of our pre-training objectives even in SOTA large-scale models.

\end{document}